\documentclass[runningheads]{llncs}
\usepackage[english]{babel}

\usepackage{lmodern}
\usepackage[T1]{fontenc}

\usepackage[letterpaper,top=2cm,bottom=2cm,left=3cm,right=3cm,marginparwidth=1.75cm]{geometry}

\usepackage{amsmath}
\usepackage{amssymb}
\usepackage[capitalize]{cleveref}
\usepackage{graphicx}
\usepackage{float}
\usepackage{wrapfig}
\usepackage{algorithm}
\usepackage{algpseudocodex}
\AtBeginEnvironment{algorithmic}{\small}

\usepackage{subcaption}
\usepackage{xspace}
\usepackage{adjustbox}
\usepackage{comment}
\usepackage{booktabs}
\usepackage{tabularx}
\usepackage{makecell}
\usepackage{tikz}
\usetikzlibrary{positioning}
\usepackage{array}
\usepackage{placeins}

\newcommand{\speclang}{\textsc{SIGMA-Spec}\xspace}
\newcommand{\approach}{\textsc{SIGMA}\xspace}

\newcommand{\eval}{\mathcal{E}}
\newcommand{\feat}{\mathcal{F}}

\newcommand{\intervals}{\mathcal{I}}
\newcommand{\ml}{\mathcal{ML}}

\newcommand{\seqnr}{\operatorname{nr}}
\newcommand{\reals}{\ensuremath{\mathbb{R}}}
\newcommand{\signals}{\mathcal{S}}

\newcommand{\trace}{\pi}

\newcommand{\sval}{\alpha} 
\newcommand{\fval}{\zeta} 

\title{Formal Specifications with\\
Data-Driven Signal Analysis%
\thanks{This work was supported by the Austrian Research
Promotion Agency (FFG) under project FATE
(project no.~894789).}}

\author{
Roderick Bloem\inst{1},
Benedikt Maderbacher\inst{2},
Cristinel Mateis\inst{3},
Dejan Ni\v{c}kovi\'{c}\inst{3}, \\
Sergii Tkachov\inst{4},
Bj\"orn Ziegler\inst{4}
}
\authorrunning{R. Bloem et al.}
\institute{Graz University of Technology, Graz, Austria\\ \email{roderick.bloem@tugraz.at} \and
CISPA Helmholtz Center for Information Security, Saarbrücken, Germany\\
\email{benedikt.maderbacher@cispa.de} \and
AIT Austrian Institute of Technology, Vienna, Austria\\
\email{\{cristinel.mateis,dejan.nickovic\}@ait.ac.at} \and
Infineon Technologies Austria AG, Villach, Austria\\
\email{\{sergii.tkachov,bjoern.ziegler\}@infineon.com}
}

\begin{document}
\maketitle
\begin{abstract}
Developing analog and hybrid systems requires substantial domain knowledge that is hard to articulate in a formal specification; without additional effort, this domain knowledge is also inaccessible to machine learning methods. We propose a hybrid approach for analyzing and monitoring real-valued signals that integrates specification-based and data-driven techniques. We use regular expressions with interval operations to identify  regions of interest in a signal, classical signal processing methods to extract semantically meaningful features, and semi-supervised machine learning to infer complex relationships that distinguish correct from faulty behaviors. The relationships learned from normal behavior data can automatically detect incorrect behavior in the same system or variants thereof. We evaluate the method on a voltage regulator, demonstrating its effectiveness in detecting undesired system behaviors.
\end{abstract}

\section{Introduction}
\label{sec:intro}

The design of analog and hybrid systems is challenging due to the tight coupling between discrete-event logic and continuous-time analog dynamics. Ensuring correctness in these systems is essential, especially for safety-critical applications where failures can have severe consequences. Formal verification methods provide rigorous guarantees, but often struggle to scale with system complexity. As a more pragmatic alternative, monitoring and evaluation of individual system behaviors against formal specifications enables an efficient, yet systematic analysis of the underlying system. Formal specification languages, such as SystemVerilog Assertions (SVA), Property Specification Language (PSL), Signal Temporal Logic (STL) \cite{DBLP:conf/formats/MalerN04} and Timed Regular Expressions (TRE) \cite{DBLP:journals/jacm/AsarinCM02}, have proven effective for this purpose. These languages allow rigorous specification of intended behaviors, facilitating the detection of property violations. Moreover, they enable pattern matching within system traces to identify critical events or anomalies. By combining formal specifications with trace evaluation, engineers can gain valuable insights into system behavior, while balancing rigor and scalability.

Nevertheless, nominally meeting the specification in an analog or hybrid system in terms of bandwidth, gain, power consumption and other desired properties, does not guarantee that the system is without flaws. Consider for instance an analog integrated circuit (IC) component with small oscillations. Such oscillations may be within the nominal specification limits and hence  not detected by the commonly used testing tools. However, they can still significantly affect the long-term reliability of the circuit. While being able to visually distinguish between benign and harmful oscillations, design experts may struggle to give their precise mathematical characterizations. It is therefore sometimes challenging to use formal specification languages alone to capture subtle anomalies in the observed system behavior.   

In this paper, we introduce  \approach, an approach for \emph{signal measurement and analysis} that integrates formal specifications with data-driven anomaly detection. At the core of \approach is \speclang, a formal specification language for expressive signal measurement and analysis. Building upon timed regular expressions (TRE) \cite{DBLP:journals/jacm/AsarinCM02,DBLP:conf/cav/FerrereMNU15}, \speclang extends their expressiveness with a flexible and powerful signal-processing layer inspired by stream-monitoring languages. Leveraging \speclang as an advanced feature extraction tool, we identify relevant signal segments and perform targeted measurements, effectively reducing the complexity of high-dimensional time series data into a concise set of meaningful parameters. These extracted features then serve as inputs for machine learning algorithms, enabling the detection of subtle and complex anomalies with improved accuracy. In this way, \approach combines explicit domain knowledge encoded through formal specifications of characteristic behavioral patterns with implicit knowledge learned from data to distinguish normal from anomalous system behaviors, yielding a powerful methodology for anomaly detection in analog and hybrid systems. We implemented our approach in a prototype tool and demonstrate its effectiveness on a voltage regulator case study.

\section{Related Work}
\label{sec:related}

Logics such as Signal Temporal Logic \cite{DBLP:conf/formats/MalerN04} extend temporal logic with the ability to specify constraints on signal values. Ferrere et al.\ \cite{DBLP:conf/cav/FerrereMNU15} instead consider regular expressions and provide a more expressive set of signal measures together with the accompanying algorithms.

We extend this work in several ways. First, we note that for runtime verification, we always have traces consisting of a finite number of samples, possibly sampled at irregular intervals. This allows us to avoid expensive zone-based algorithms and employ standard techniques for matching regular expressions. Second, we use formal specifications not only to identify semantically meaningful intervals within signals, but on these intervals, we also apply signal-processing operations and extract features that capture relevant behavioral characteristics. We then apply machine-learning techniques to these specification-derived features to distinguish correct from suspicious traces. We thus combine domain-guided symbolic localization with data-driven anomaly detection, reducing the burden on the verification engineer of specifying correctness explicitly while still enabling the use of domain knowledge.

Matrix Profile methods support efficient time-series subsequence mining, e.g., motif and discord discovery~\cite{DBLP:conf/icdm/YehZUBDDSMK16}.
They are complementary to \speclang: while Matrix Profile discovers recurring or unusual patterns directly from raw time series, \speclang first uses symbolic specifications to localize semantically relevant intervals, such as post-perturbation stabilization regimes, before applying feature operators and learned predicates.

The use of machine-learning methods is fairly standard for detecting anomalies in signals across various domains \cite{DBLP:conf/kdd/HundmanCLCS18,DBLP:conf/bigdataconf/AhmadSNK20,Esmaei23}. These approaches typically use machine learning in isolation, without logical representations. This means that domain knowledge has to be included into the machine-learning algorithm in an ad hoc manner. 

Kong et al.\ \cite{DBLP:journals/tac/KongJB17} also study a combination of formal specification and machine learning. They define a language of parameterized specifications and use anomaly learning and anomaly detection to learn formulas and parameter values that distinguish good traces from bad ones.

Several papers study quantitative versions of logics that incorporate robustness measures \cite{DBLP:journals/tcs/FainekosP09,DBLP:conf/formats/BakhirkinFMU17,DBLP:journals/fmsd/DeshmukhDGJJS17}. These approaches focus on quantitative semantics for logical satisfaction rather than on signal-processing-based feature extraction and learned classification of real-valued signals. In our approach, robustness is not modeled explicitly in the formalism, but is instead handled through the confidence estimates provided by the machine-learning methods.

\section{\speclang Specification Language}
\label{sec:sigma}

\subsection{Language}
\begin{figure*}[tbp]
    \centering
    \includegraphics[width=1.0\linewidth]{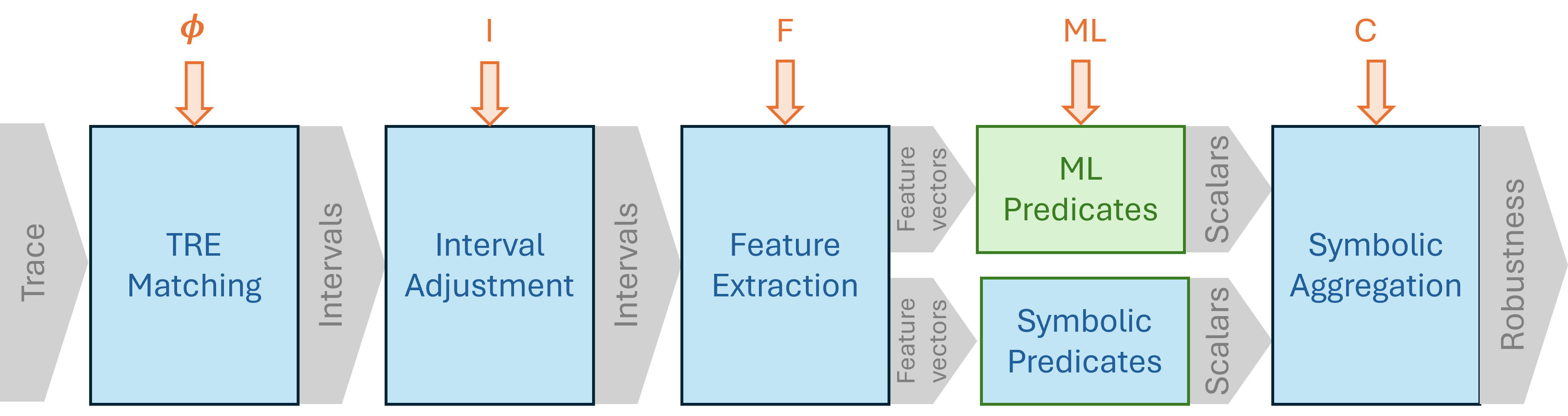}
    \caption{\approach pipeline}
    \label{fig:pipeline}
\end{figure*}

\speclang is a stream processing language to specify and process intervals of a trace. It is illustrated in \cref{fig:pipeline}: First, we match intervals using regular expressions over events. We can then adjust the intervals by shifting their boundaries. Next, features are extracted and a Machine Learning step  assigns a robustness score. Finally, the robustness scores of the intervals are aggregated in an overall robustness score for a trace.

Formally, we assume that a measurement trace contains samples of a set of real-valued \emph{signals} $\signals$. (This is easily extended to Boolean signals.) We also assume that we have a set $\feat$ of real-valued \emph{features} that are computed as part of the processing pipeline. 
Finally, we have a set $\ml$ of machine learned classifiers that each assign a robustness score in $\reals$. A \emph{valuation} is a  function $\sval: \signals \rightarrow \reals$ and $\eval$ is the set of valuations.  Finally, a \emph{trace} is a finite sequence $\trace \in (\reals, \eval)^*$ over timestamps and valuations such that the timestamps are strictly increasing.
Note that we only have a finite number of samples that may be taken at irregular intervals.

An \emph{interval record} is a tuple $\rho = (\trace, t, t', \fval)$ where $\trace = (t_1,\sval,\dots,t_n, \sval_n)$ is a trace, and $t,t' \in \reals$ are the beginning and end times. 
Each interval record is associated with a partial function $\fval : \feat \rightharpoonup \reals$ assigning values to features. 
The start and end times  $t, t'$ corresponds to a sub-trace $\trace'= (t_i,\sval_i,\dots,t_j, \sval_j)$ of $\trace$, where $t_i$ ($t_j$) is the smallest (largest) timestep that is not smaller (larger) than $t$ ($t'$, resp.).
We assume that our feature set includes a distinguished feature $\seqnr$, which we use to order the matched sequences.
We denote the set of intervals by $\intervals$.

Trace processing consists of five stages, which we briefly summarize here. (1) We use Timed Regular Expressions (TRE, see below) to extract  interval records from traces. (2) Intervals can be adjusted, for instance, we can drop a suffix. (3) new features are extracted. (4) A machine learning algorithm computes a robustness score for the interval. Finally, (5) the robustness scores of the intervals are aggregated into an overall score for the trace.

\begin{wrapalgorithm}{R}{0.50\textwidth}
\caption{Evaluation of a specification $\phi$ on a trace $\pi$.}
\label{alg:evalspec}
\begin{algorithmic}
\State $\varphi, P, c \gets \phi$ \Comment{TRE $\varphi$, pipeline stages $P$, aggregator $c$}
\State $M \gets \mathcal{M}(\varphi, \pi, 0, t_{max})$ \Comment{generate  interval records}
\For{$p \in P$}
    \State $M' \gets \emptyset$
    \For{$(\trace, t, t', \fval) \in M$}
        \If{$p$ is an interval adjustment}
            \State $M' \gets M' \cup p(\trace, t, t', \fval)$
        \EndIf
        \If{$p$ is a feature operation}
            \State $M' \gets M' \cup \{ (\trace,t,t', \fval[p \mapsto p(\trace, t, t', \fval)] \}$
        \EndIf
    \EndFor
    \State $M \gets M'$
\EndFor
\State $r \gets c(M)$
\end{algorithmic}
\end{wrapalgorithm}

If $\phi$ denotes a TRE, $I$ denotes an interval adjustment operator, $f$ denotes the extraction of a feature $f \in \feat$, $m$ denotes the extraction of a robustness score using a machine-learning algorithm $m \in \ml$, and $c$ denotes an aggregation function, then a \speclang specification has the form
\begin{align*}
spec    &::= \varphi \ (\triangleright\ process)^* \ \blacktriangleright\ c\\
process &::= I \mid f \mid m.
\end{align*}

Algorithm~\ref{alg:evalspec} shows the evaluation of a specification. In the following we describe the individual components of \speclang in more detail.

\subsection{Interval Recognition}

We specify interesting intervals with \emph{timed regular expressions} (TREs) \cite{DBLP:conf/formats/UlusFAM14,DBLP:conf/cav/FerrereMNU15}. TREs are defined over \emph{propositions} matched over intervals of time. We take propositions to be either Boolean combinations of numeric predicates of the form $x \bowtie c$ (with $x$ a real-valued signal variable, $\bowtie \; \in \{ <, \leq, =, \geq, >\}$, and $c$ a constant), or \emph{events} defining \emph{rising edge} $\uparrow p$ ($p$ goes from false to true) and \emph{falling edge} $\downarrow p$ of a predicate $p$.

The TRE syntax follows the grammar
$$
    \varphi ::= \epsilon \mid p \mid \varphi_1 \cdot \varphi_2 \mid \varphi_1 \cup \varphi_2 \mid \varphi_1 \cap \varphi_2 \mid \varphi^{*} \mid \langle \varphi \rangle_I,
$$
where $p$ is a proposition and $I \subset \reals$ is an interval. This syntax defines a variant of classical regular expressions, extended with an \emph{intersection} $\cap$ and a \emph{time restriction} $\langle \cdot \rangle_{I}$ operators. 

Let  $\trace = (t_1, \sval_1), \dots,(t_n, \sval_n)$ be a trace and let $\rho = (\trace, t, t')$ be an interval. TREs follow the usual semantics of regular expressions on words, being matched by intervals $\rho' = (\trace, u, u')$, where $u\geq t$ and $u' \leq t'$. (We denote this by $\rho' \models \varphi$.) The $\langle \varphi \rangle_I$ operator is matched by any interval $(\trace, u, u')$ where $u = t_i$ for some $i$, there is a $d \in I$ such that $u' = u + d$, and $(\trace, u, u') \models \varphi$.

Following standard greedy matching semantics, the \emph{first match} of a TRE $\varphi$ in $\rho$ is $(\trace, u, u')$, where $u$ is minimal and $u'$ is maximal, i.e., there is no $v<u$ and $v'$ with $(\trace, v, v') \models \varphi$ and there is no $v'>u'$ so that $(\trace, u, v') \models \varphi$. 
The \emph{extended match set} $\mathcal{M}(\varphi, (\trace, t, t'))$ is recursively defined as the set consisting of the union of the singleton set containing the interval record $(\trace, u, u', \fval)$ of the first match and $\mathcal{M}(\varphi, (\trace, u', t'))$, where (with some abuse of notation) $\fval$ assigns $i$ to feature $\seqnr$ of the $i$-th match and is undefined everywhere else.

Note that for a sampling semantics, the complexity of matching regular expressions is equal to that of matching regular expressions on words. 

\subsection{Interval Adjustment}
In addition to the initial matching using TREs \speclang supports operators  that modify the matched intervals.
An interval operation is a function $\intervals \rightarrow 2^{\intervals}$ that maps an interval record to a set of interval records, without changing features. 
\speclang includes the following interval operations, where $\rho = (\trace, t, t', \fval)$, $I \in \reals$ is a duration, $b,e \in [0,1]$ with $r_1<r_2$, $l,s \in \mathbb{N}$, and $p: \fval \to \mathbb{B}$:
\begin{align*}
    \text{prefix}_I(\rho) &= \{(\trace, t, \min(t+I, t'), \fval)\},\\
    \text{postfix}_{I}(\rho) &= \{(\trace, \max(t, t'-I), t', \fval)\},\\
    \text{slice}_{b,e}(\rho) &= \{(\trace, t+b(t'-t), t+e(t'-t),\fval)\} \\
    \text{filter}_p(\rho) &= \{\rho\}  \text{ if }  p(\fval) \text{ else } \emptyset\\
    \text{linefit}_{l,s}. &
\end{align*}
The function linefit splits the interval $\rho$ into $l$ sub-intervals using line fitting \cite{ozay2016exact}. It finds a piece-wise linear continuous function consisting of $l$ line segments that best approximate the interval and returns the interval for the line segment $s$.

\subsection{Feature Extraction}

Feature operations are functions $f: \intervals \to \mathbb{R}$.
When evaluating the pipeline the feature is computed for each interval record and the result is added to feature valuation $\fval$.

\speclang supports a  variety of feature operations. We will discuss some examples. A trace can be sampled at irregular intervals which makes some features hard to compute. These features are computed using a linear approximation between samples.

\speclang includes the following feature operations:
(1) The features \textbf{minimum} and \textbf{maximum} and \textbf{average} are computed directly from the trace.  
(2) The \textbf{linear fit slope} and \textbf{linear fit MSE} are the slope and mean square error of the linear approximation of the signal (see the previous section), resp.
(3) Several spectral features: the \textbf{spectral peak-to-mean ratio} is the ratio between the maximum and the mean spectral amplitude, measuring the dominance of a single frequency component over the overall spectral content. The  \textbf{spectral centroid} and \textbf{spectral bandwidth} give the amplitude-weighted average frequency and standard deviation, resp., of the signal spectrum, providing a compact descriptor of where and how strongly  spectral energy is concentrated.
(4) \textbf{Symbolic predicates} are functions that compute a robustness score from one or more existing features indicating how anomalous they are. A classic example would be the distance to a threshold.

\subsection{ML-based Anomaly Scoring Functions}
\label{sec:ml-scoring}

In addition to feature operations with fixed symbolic definitions, \speclang allows externally trained anomaly detectors to be applied to interval records.
Let $(\pi, t, t', \fval)$ be an interval record, where $\fval$ is the feature valuation computed for the interval $i=[t,t']$ of trace $\pi$.
An ML-based anomaly detector provides an anomaly scoring function $a_\theta:\mathcal{V}\rightarrow\mathbb{R}$, where $\mathcal{V}$ denotes the space of feature valuations and $\theta$ the parameters of the ML model learned during training.
Evaluating $a_\theta$ on the current feature valuation yields a real-valued anomaly score. 
The score can be stored as an additional feature by extending $\fval$ to $\fval'$ with
\[
    \fval'(a_\theta) = a_\theta(\fval).
\]
Thus, learned anomaly scores can be used by interval conditions in the same way as other numeric features, for instance by comparing them against a threshold $\tau_a$.

This subsection only defines the interface between \speclang specifications and learned scores.
The motivation and details of concrete ML-based predicates are discussed in \cref{sec:ml-predicates}.

\subsection{Aggregation of Interval Streams}
We often want to combine the results from different intervals into an overall score for a trace.
To facilitate this \speclang includes a small sub-language to describe this.
Its grammar is defined as:
$$
c ::= \text{max}_f \mid \text{min}_f \mid c + c \mid c - c,
$$
where $\text{max}_f$ evaluates to the maximum of the feature $f$ among all intervals and $\text{min}_f$ computes the respective minimum.
Addition and subtraction can be used to combine the results computed from different features.

A numerical anomaly score obtained in this way can be used in multiple ways.
Given a set of traces, the score can be used to sort them by their level of anomaly. A verification engineer can then limit manual inspection to a small subset of traces to  identify potential  problems more quickly.
Alternatively, a Boolean predicate can be defined to classify individual traces, by defining a threshold for when a trace should be considered anomalous. If the final score is based on an ML-based predicate its threshold $\tau$ is a natural choice for such a predicate.

\subsection{Example}
To help with understanding \speclang we provide the following example:
\[
\begin{split}
\Phi = &(\uparrow_{s>0.3} \cdot (\neg\downarrow_{s>0.3})^* \cdot \downarrow_{s>0.3}) \operatorname{\triangleright}
         \text{filter}_{\seqnr=2 \lor \seqnr=3} \\
         &\operatorname{\triangleright} \text{postfix}_{(t'-t)*0.3}
        \operatorname{\triangleright} \text{slope}_s \operatorname{\blacktriangleright} \text{max}_{\text{slope}_s} - \text{min}_{\text{slope}_s}
        \end{split}
\]
The TRE pattern defines intervals delimited by rise and fall events of $s>0.3$.
The found intervals are altered with the interval operations filter and postfix.
The filter uses the sequence number $n$ to select only the second and third match.
Postfix then selects the last $30\%$ time units of each interval.
As feature we use the slope of the linear approximation. 
Finally the aggregator computes the difference between the maximum and minimum value of that feature among all selected intervals.

\begin{wrapfigure}{L}{0.5\linewidth}
    \centering
    \includegraphics[width=\linewidth]{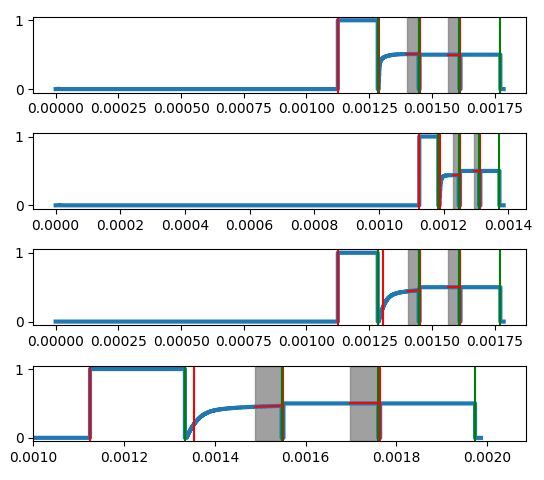}
    \caption{Quantitative analysis of analog simulations.}
    \label{fig:quantitative}
\end{wrapfigure}

Figure~\ref{fig:quantitative} shows how this specification is applied to four simulations of an oscillator. The signal $s$ is the blue line. Rise and fall events are shown as red and green vertical lines respectively. The gray shaded areas are the intervals remaining after the interval operations. The slope feature is shown by a red line of the linear approximations in each interval.

\section{ML-based Predicates in \speclang}
\label{sec:ml-predicates}

Symbolic predicates are effective when nominal signal behavior can be expressed by explicit arithmetic constraints over measurements.
In analog-design workflows, however, subtle effects such as sustained oscillations or abnormal stabilization patterns are often easier for experts to label from examples than to characterize completely by hand.

The scoring interface introduced in Section~\ref{sec:ml-scoring} allows such examples to be used for learned interval predicates.
Given a trace $\pi$ and an interval $i$, the preceding \speclang analysis localizes the relevant signal segment and computes a feature valuation $\fval_{\pi,i}$.
We write
\[
    \mathbf{x}_{\pi,i}=(f_1(\pi,i),\ldots,f_d(\pi,i)) \in \mathbb{R}^d
\]
for its vector representation.
A learned predicate is defined by an anomaly scoring function $a_\theta:\mathbb{R}^d\to\mathbb{R}$ and a threshold $\tau_a$:
\[
    \mathsf{anom}_{\theta,\tau_a}(\pi,i) =
    \bigl(a_\theta(\mathbf{x}_{\pi,i}) > \tau_a\bigr).
\]
Thus, the specification determines where to look and which measurements to compute, while the learned predicate decides whether the resulting feature vector is consistent with nominal behavior.

We focus on a semi-supervised one-class setting: only samples labeled as normal are used to fit the learned predicate, while anomalous samples are reserved for feature selection and evaluation.
This reflects the industrial setting in which faults are scarce and expensive to label.
The choice of the learning algorithm remains data-dependent. \speclang provides the specification-derived representation, but the learned predicate must still be selected and validated for the feature relationships present in the dataset.

In the evaluation, we instantiate learned predicates with autoencoders and One-Class SVMs.
Autoencoders provide reconstruction-based anomaly scores, using
\[
    a_{\mathrm{AE}}(\mathbf{x}) =
    \left\lVert \mathbf{x} - AE_\theta(\mathbf{x}) \right\rVert_2^2,
\]
and have been used for nonlinear anomaly detection in time-series and telemetry data~\cite{DBLP:conf/pricai/SakuradaY14}.
One-Class SVMs estimate the support of the normal data distribution for novelty detection~\cite{DBLP:conf/nips/ScholkopfWSSP99}.
Given their decision function $g_{\mathrm{OCSVM}}$, we use $a_{\mathrm{OCSVM}}(\mathbf{x})=-g_{\mathrm{OCSVM}}(\mathbf{x})$ as anomaly score, so that larger values indicate stronger deviation from the estimated normal support and can be thresholded like the AE reconstruction error.
Comparing the two on the same \speclang-derived feature vectors evaluates the sensitivity of the learned-predicate component to the chosen one-class model. It does not test the usefulness of \speclang itself, since localization and feature extraction are shared.

\section{Evaluation}
\label{sec:evaluation}

\subsection{Case Study: Voltage Regulator}
\label{sec:case-study-data}

\begin{figure}[t]
\centering
\begin{tikzpicture}[
block/.style={
draw,
rounded corners,
minimum width=1.55cm,
minimum height=0.65cm,
align=center
},
smallblock/.style={
draw,
rounded corners,
minimum width=1.15cm,
minimum height=0.5cm,
align=center
},
>=latex,
node distance=0.75cm
]

\node (vin) {$V_{\mathrm{in}}$};
\node[smallblock, right=0.45cm of vin] (cin) {$C_{\mathrm{in}}$};
\node[block, right=0.55cm of cin] (core) {pass\\device};
\node[smallblock, right=0.55cm of core] (cout) {$C_{\mathrm{out}}$};
\node[right=0.45cm of cout] (vout) {$V_{\mathrm{out}}$};

\draw[->] (vin) -- (cin);
\draw[->] (cin) -- (core);
\draw[->] (core) -- (cout);
\draw[->] (cout) -- (vout);

\node[smallblock, below=0.6cm of cout] (load) {load};
\draw[->] (cout) -- node[right] {$I_{\mathrm{load}}$} (load);

\node[smallblock, below=0.8cm of core] (ctrl) {control /\\feedback};
\node[left=0.45cm of ctrl] (vref) {$V_{\mathrm{ref}}$};

\draw[->] (vref) -- (ctrl);
\draw[->] (ctrl) -- (core);
\draw[->] (load.south) |- (ctrl.east);

\end{tikzpicture}
\caption{Simplified closed-loop view of the voltage-regulator case study.
The regulator compares feedback from $V_{\mathrm{out}}$ with a reference
value $V_{\mathrm{ref}}$ and controls the pass device to compensate for input
voltage and load-current perturbations.}
\label{fig:vr-overview}
\end{figure}
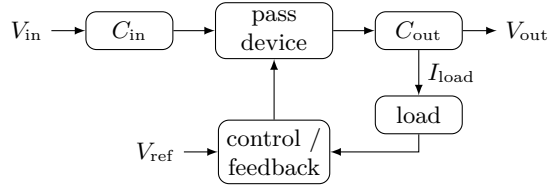

We introduce as our case study the model-based testing of a voltage regulator. A voltage regulator is an analog circuit that converts a possibly varying input supply voltage $V_{\mathrm{in}}$ into a regulated output voltage $V_{\mathrm{out}}$ close to a prescribed reference value $V_{\mathrm{ref}}$.
In a typical closed-loop regulator, the output voltage is sensed through a feedback path and compared with the reference voltage; the resulting control signal drives a pass device so as to compensate for variations in the input voltage and in the load current.
Capacitors are used to filter the input and output nodes and influence the transient response and stability of the regulator~\cite{ti1999ldo}.
Figure~\ref{fig:vr-overview} shows the high-level structure of a voltage regulator.

We assume a model-based testing setting, in which we have both a behavioral model of the voltage regulator, as well as its transistor-level design. In this approach, the behavioral model serves as the basis for test generation and for defining test oracles for the circuit-level design. We treat the voltage regulator as a black-box system, interacting with it solely through simulation while observing and analyzing its input and output signals.
The analyzed signal is the sampled output voltage $V_{\mathrm{out}}$.
Perturbations are induced by changes in the input voltage and/or load current.
A typical response consists of an initial transient, possibly including an overshoot or undershoot, followed by a stabilization regime in which the output voltage returns close to the target reference value. While the initial transient, the overshoots and the undershoots admit efficient symbolic analysis, the analysis of the stabilization regime remains a hard challenge to formalize with specifications.
In this case study, we focus on the detection of anomalies in the stabilization regime -- its sustained oscillations, excessive residual periodicity, or abnormal noise patterns after the transient may indicate an incorrect design configuration or deteriorated component behavior.

\subsection{Research Questions}
We seek the answers to the following questions:
\begin{description}
    \item[RQ1] How accurate is \approach's in-domain anomaly detection? 

    \item[RQ2] How well does \approach trained on signals generated from a behavioral model generalize to:
    \emph{(G1)} multi-parameter traces of the same model,  
    \emph{(G2)} the same model with different parameters, and
    \emph{(G3)} transistor-level implementation of the same model? 
    
    \item[RQ3] Do learned predicates provide an advantage over purely symbolic predicates? 
    
    \item[RQ4] How sensitive is \approach to the choice of the machine learning algorithm used to infer the learned predicate?
\end{description}

\subsection{Experimental Setup}

\subsubsection{Voltage Regulator Models}

We evaluate the case study with two models: a behavioral model of the voltage regulator and its transistor-level design.

\emph{InfineonSpice behavioral modelling:}
InfineonSpice is an analog SPICE simulation tool for behavioral circuit simulation with SPICE models, virtual reference designs, and transient-analysis support~\cite{infineonspice}.
In our experiments, InfineonSpice enables inexpensive trace generation that we use to produce the data for training monitors for detecting anomalies.

We consider three VR designs with target reference voltages 5.0\,V, 3.3\,V, and 1.5\,V.
For each design, we vary the parameters listed in Table~\ref{tab:vr-parameters}.
The parameter variations serve two purposes.
First, variations of internal design parameters such as $p_{\mathrm{cin}}$, $p_{\mathrm{cout}}$, and the PID-controller parameters $p_{\mathrm{kd}}$, $p_{\mathrm{ki}}$, and $p_{\mathrm{kp}}$ are used to emulate non-nominal or deteriorated behavior.
Second, variations of operating and stimulus parameters such as $p_{\mathrm{vin}}$, $p_{\mathrm{vin\_delta}}$, and $p_{\mathrm{load\_step}}$ explore different input-voltage and load-current perturbations.
We also vary $p_{\mathrm{vref}}$ to study cases in which the regulator is operated with a reference value different from the nominal value of the model design.

\begin{wraptable}{R}{0.48\textwidth}
\centering
\caption{VR parameters varied in the InfineonSpice simulations.}
\label{tab:vr-parameters}
\begin{adjustbox}{max width=\linewidth}
\begin{tabular}{l l}
\toprule
\textbf{Parameter} & \textbf{Meaning} \\
\midrule
$p_{\mathrm{vref}}$ & prescribed reference voltage \\
$p_{\mathrm{cout}}$ & output capacitor \\
$p_{\mathrm{vin}}$ & input voltage \\
$p_{\mathrm{cin}}$ & input capacitor \\
$p_{\mathrm{kd}}$ & PID derivative gain \\
$p_{\mathrm{ki}}$ & PID integral gain \\
$p_{\mathrm{kp}}$ & PID proportional gain \\
$p_{\mathrm{load\_step}}$ & load-current increase \\
$p_{\mathrm{vin\_delta}}$ & input-voltage increase \\
\bottomrule
\end{tabular}
\end{adjustbox}
\end{wraptable}

For each design, we generate single-parameter sweeps by varying one parameter at a time up to $\pm 50\%$ from its nominal value in 5\% steps, plus one fully nominal trace.
This yields 181 traces per design.
Additional multi-parameter traces are generated by randomly sampling all parameters from their ranges until obtaining 5000 traces per design.

\begin{figure*}[t]
    \centering
    \begin{subfigure}[t]{0.48\textwidth}
        \centering
        \includegraphics[width=\linewidth]{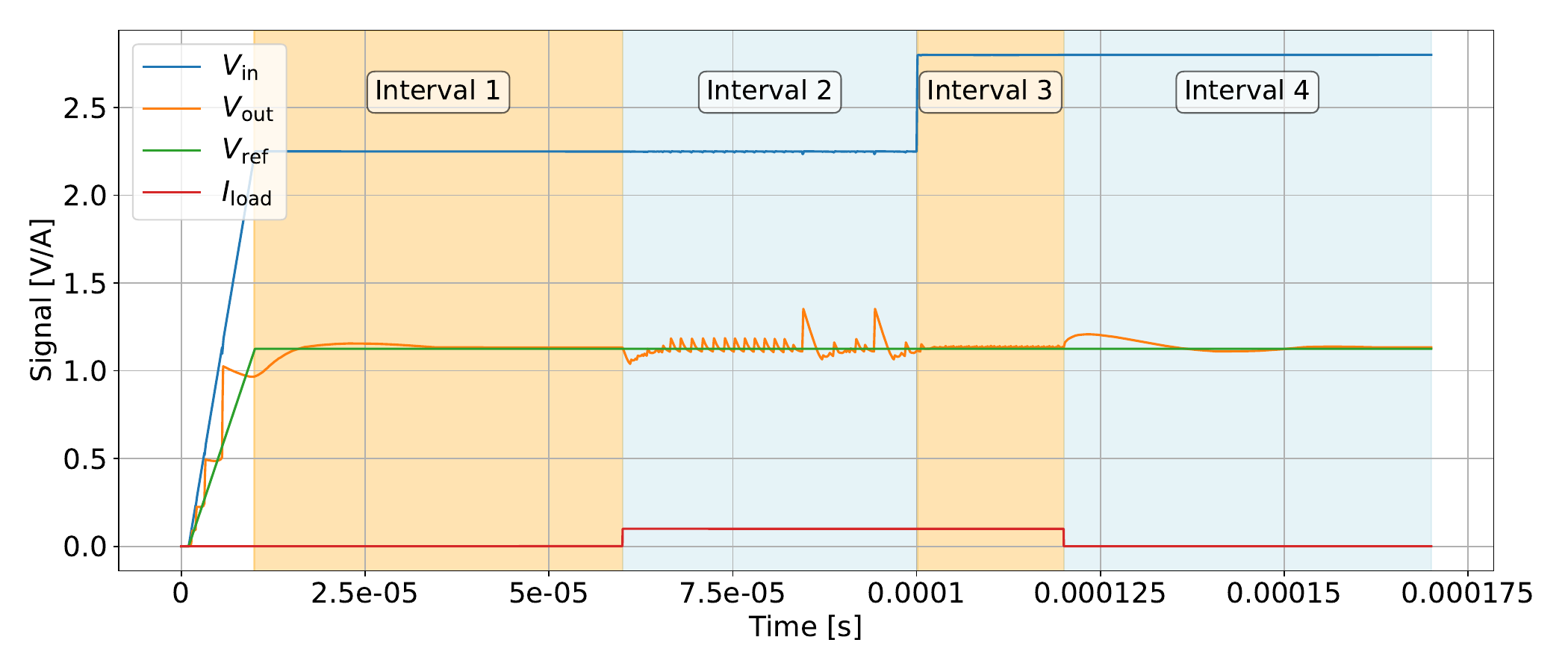}
        \caption{InfineonSpice trace. \speclang detects four perturbation
        intervals. The first, third, and fourth intervals are labeled normal;
        the second interval is labeled anomalous due to a dominant repeating
        triangular pattern followed by a shorter repeating pattern of more
        complex shape.}
        \label{fig:example-trace-infineonspice}
    \end{subfigure}
    \hfill
    \begin{subfigure}[t]{0.48\textwidth}
        \centering
        \includegraphics[width=\linewidth]{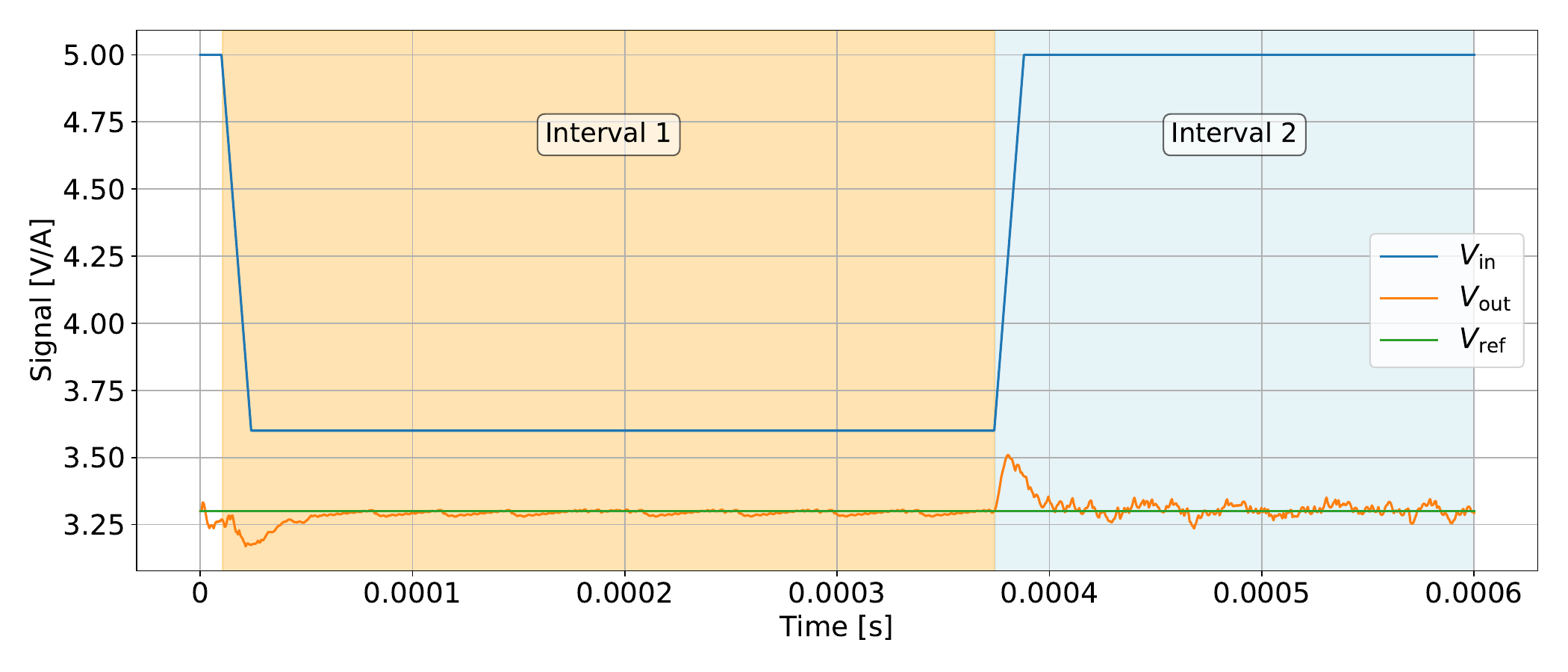}
        \caption{Transistor-level trace. \speclang detects two perturbation
        intervals. The first interval is labeled anomalous due to a dominant
        repeating triangular pattern, whereas the second interval is labeled
        normal since it shows low-amplitude noise rather than a repeating
        oscillatory pattern.}
        \label{fig:example-trace-transistor}
    \end{subfigure}
    \caption{Representative full traces from the two simulation environments
    used in the case study. The colored regions indicate perturbation
    intervals detected by \speclang and illustrate the type of signal behavior
    later analyzed at the level of stabilization portions.}
    \label{fig:full-trace-examples}
\end{figure*}

For each InfineonSpice simulation, the input voltage and load current are varied sequentially:
(i) an initial ramp from 0 to $p_{\mathrm{vin}}$, which is discarded from the dataset;
(ii) a load-current increase by $p_{\mathrm{load\_step}}$;
(iii) an input-voltage increase by $p_{\mathrm{vin\_delta}}$; and
(iv) a return of the load current to its original value.
This protocol yields four perturbation intervals per trace after \speclang segmentation.
Figure~\ref{fig:full-trace-examples}(a) shows a representative InfineonSpice trace with \speclang segmentation.

\emph{Transistor-level simulation:}
We use a proprietary closed-source simulator to implement a transistor-level design of a VR and generate the accurate VR behavior.
These traces are more representative of the physical circuit behavior but are significantly more expensive to obtain: one simulation takes approximately 10 hours on average and up to 18 hours in the worst case.
Consequently, transistor-level traces are used only for evaluation, not for training.

The transistor-level dataset consists of 33 traces for a VR model designed for a target reference voltage of 3.3\,V.
Simulations were run for three process corners: fast, nominal, and slow.
The temperature was fixed to $27^\circ$C.
All VR design parameters were kept at nominal values except the load current, which was varied from 90\,mA to 190\,mA in steps of 10\,mA. yielding 11 traces per process corner and 33 traces in total.
During each simulation, the input voltage was stepped from 5\,V to 3.6\,V and then back to 5\,V.
The initial constant 5\,V interval was too short for analysis and is discarded.
Therefore, each transistor-level trace yields two perturbation intervals after \speclang segmentation.
A representative transistor-level trace is shown in Figure~\ref{fig:full-trace-examples}(b).

\subsubsection{Stabilization Specification and Labeling}
\label{sec:sigma-extraction-quality}

We describe in this section how we formulate specifications that identify the stabilization intervals in signals and how we label the training data.

\begin{figure*}[t]
    \centering
    \begin{subfigure}[t]{0.32\linewidth}
        \centering
        \includegraphics[width=\linewidth]{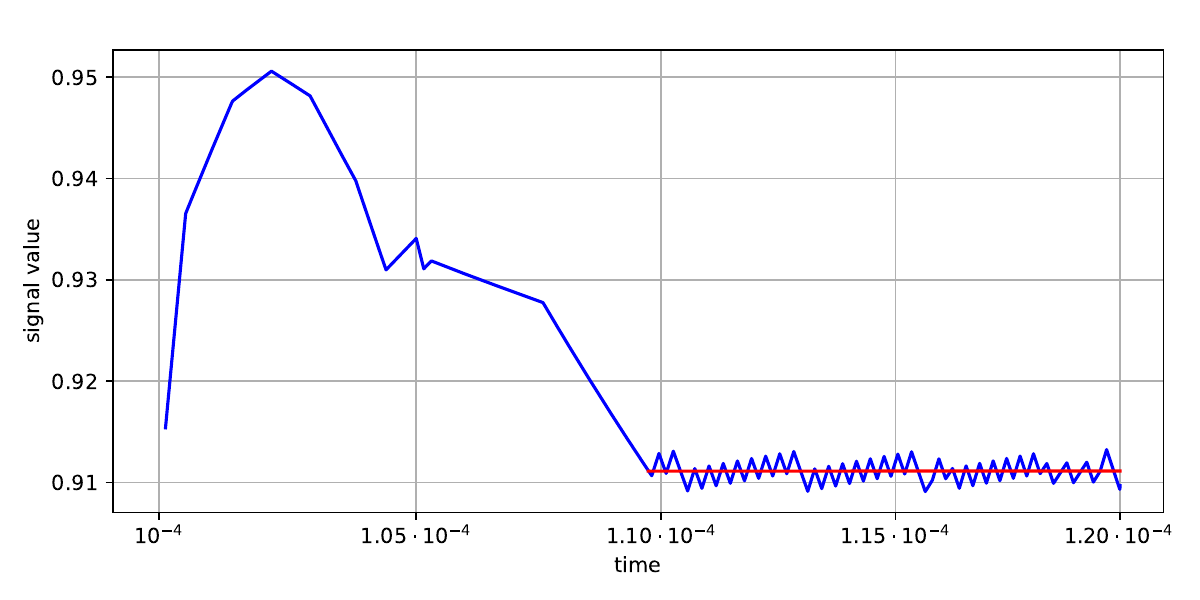}
        \caption{Valid.}
        \label{fig:valid_stab}
    \end{subfigure}
    \hfill
    \begin{subfigure}[t]{0.32\linewidth}
        \centering
        \includegraphics[width=\linewidth]{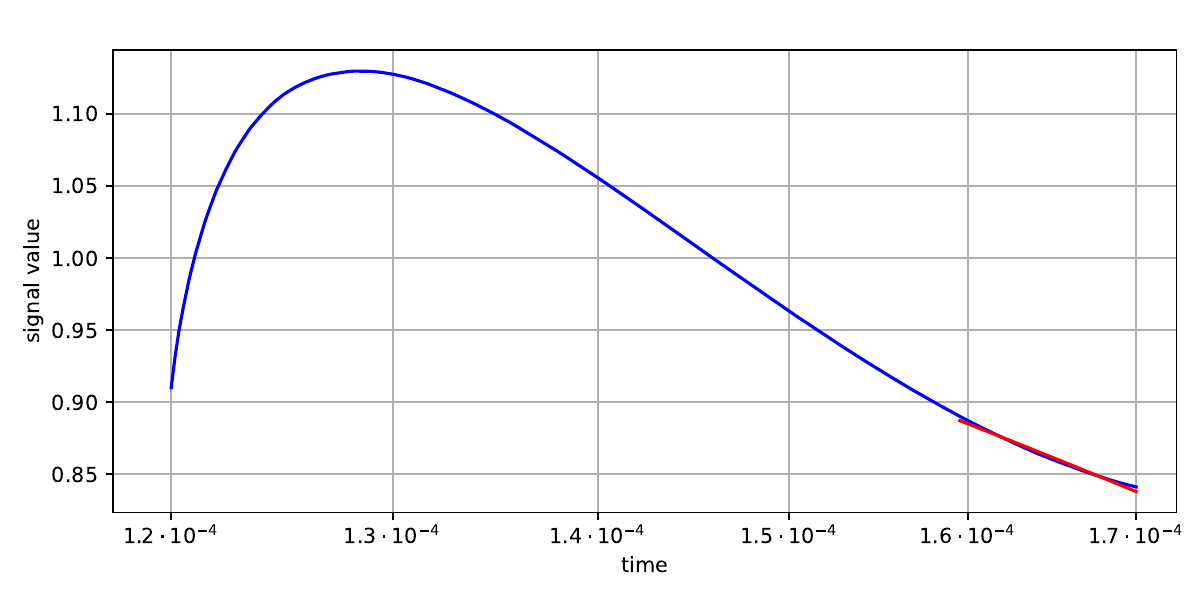}
        \caption{No stabilization.}
        \label{fig:no_stab}
    \end{subfigure}
    \hfill
    \begin{subfigure}[t]{0.32\linewidth}
        \centering
        \includegraphics[width=\linewidth]{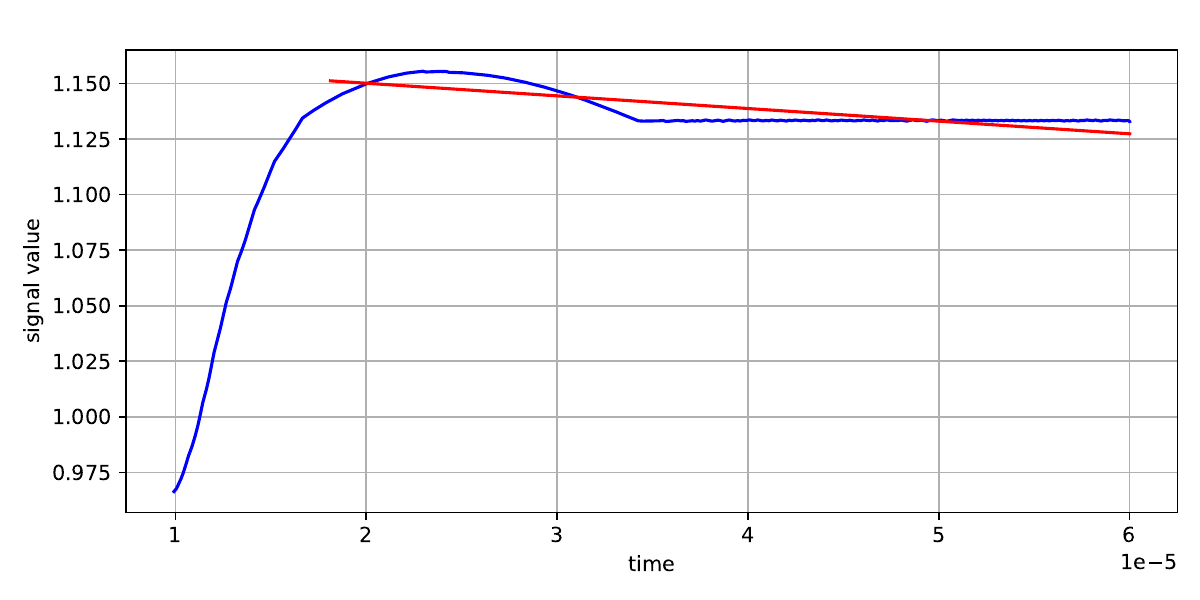}
        \caption{Incorrect fit.}
        \label{fig:missed_stab}
    \end{subfigure}
    \caption{Examples of stabilization extraction. 
    \textbf{Left:} correct stabilization detection.
    \textbf{Middle:} missing stabilization due to insufficient simulation time.
    \textbf{Right:} an inaccurate third-segment fit that still includes part of the transient.}
    \label{fig:stabilization-examples}
\end{figure*}

\emph{Specification of stabilization intervals:}
\speclang is used to identify perturbation intervals in each trace.
By construction, each InfineonSpice trace yields four perturbation intervals after discarding the initial ramp phase, while each transistor-level trace yields two usable perturbation intervals. Within each perturbation interval, a piece-wise linearization operator fits a three-segment approximation of the output voltage signal.
The third segment is intended to capture the stabilization regime.
This extraction is not always valid: the signal may not stabilize within the simulated time horizon, or the third segment may still contain part of the transient overshoot or undershoot.
Figure~\ref{fig:stabilization-examples} illustrates one valid case and two typical failure modes.

\emph{Labeling:}
A domain expert visually inspected the simulation traces and labeled the normal and the anomalous stabilization intervals in the traces. The expert labeled in total 289 traces -- 181 labeled traces were used as the training data to infer the anomaly detectors, while the remaining 108 labeled traces were used as ground truth for the evaluation of our \approach approach. We also note that in addition to the validation of the normal/anomalous labels, the expert also validated the validity of the extracted stabilization intervals.
Invalid stabilization samples were excluded from training and evaluation.

Table~\ref{tab:dataset-summary} summarizes the datasets used in the evaluation after \speclang segmentation and expert validity filtering.
We use the labeled 5.0\,V single-parameter dataset for model selection and in-domain testing. 
The remaining rows correspond to the three transfer settings used for the evaluation of RQ2.

\begin{table}[t]
\centering
\caption{Datasets after \speclang segmentation and expert validity filtering. The ``valid'' column reports valid stabilization samples labeled by domain experts.}
\label{tab:dataset-summary}
\begin{adjustbox}{max width=\linewidth}
\begin{tabular}{l l l c c}
\toprule
\textbf{Use} & \textbf{Design/source} & \textbf{Simulator} & \textbf{Traces} & \textbf{Valid ok/nok} \\
\midrule
Model selection & 5.0\,V, single-param. & InfineonSpice & 181 & 513/127 \\
In-domain test & 5.0\,V, single-param. & InfineonSpice & -- & 103/127 \\
G1 & 5.0\,V, multi-param. & InfineonSpice & 25 & 38/10 \\
G2 & 3.3\,V & InfineonSpice & 25 & 32/7 \\
G2 & 1.5\,V & InfineonSpice & 25 & 27/14 \\
G3 & 3.3\,V & transistor-level & 33 & 58/6 \\
\bottomrule
\end{tabular}
\end{adjustbox}
\end{table}

\subsubsection{Learned-Predicate Instantiation}
\label{sec:learned-predicate-inst}
All learned predicates use standardized features, with the scaler fitted only on normal training samples.
Feature subsets are selected on the labeled 5.0\,V single-parameter data using nested 5-fold cross-validation over the normal samples.
The held-out normal fold is combined with the labeled anomalous samples for testing, while the remaining normal samples are split into training and validation sets.
For the AE, we use a symmetric encoder-decoder with hidden dimensions $N{:}M{:}N$, where $N=\max(8,\lfloor d/2\rfloor)$ and $M=\max(4,\lfloor d/4\rfloor)$ for input dimension $d$.
It is trained with Adam, learning rate $10^{-3}$, batch size 128, and early stopping.
The AE threshold is set to $\mu+2\sigma$ on log-transformed validation-normal reconstruction errors.
For OCSVM, we use an RBF kernel with \texttt{gamma=scale}, $\nu=0.05$, and a threshold at the 95th percentile of validation-normal anomaly scores.
These are case-study instantiations, not generally optimal settings.

\subsection{RQ1: In-Domain Accuracy}
\label{sec:model-selection}

From each valid stabilization segment, \speclang feature operators extract time- and frequency-domain measurements such as linear-fit error, slope, offset, spectral energy, spectral flatness, and periodicity.

Table~\ref{tab:ae-models} summarizes the two AE configurations retained after feature-subset selection:
AE-R1, the best in-domain feature set, and
AE-C, a compact four-feature set with similar cross-validation performance.

\begin{table}[t]
\centering
\caption{Selected AE feature sets. Test entries are accuracy/F1(ok)/F1(nok).}
\label{tab:ae-models}
\begin{adjustbox}{max width=\linewidth}
\begin{tabular}{l l c c c}
\toprule
\textbf{ID} & \textbf{Features} & \textbf{\#} & \textbf{CV F1} & \textbf{Test} \\
\midrule
AE-R1 & \makecell[l]{slope, offset, mse, p2m, \\ bw, flat., energy, periodicity} & 8 & $0.902 \pm 0.006$ & $0.878/0.877/0.879$ \\
AE-C & slope, offset, centroid, energy & 4 & $0.888 \pm 0.013$ & $0.870/0.867/0.872$ \\
\bottomrule
\end{tabular}
\end{adjustbox}
\end{table}

Both AE configurations  
achieve balanced in-domain performance, with F1 scores around $0.87$--$0.88$ for both classes.
The compact feature set loses less than one percentage point in accuracy while using half as many features.
\textbf{This answers RQ1 positively: \approach achieves accurate in-domain anomaly detection.}

\subsection{RQ2: Generalization Analysis}
\label{sec:generalization-analysis}

We evaluate the two selected AE predicates on the three generalization scenarios G1--G3 defined in Section~\ref{sec:evaluation}.
The same trained models and thresholds are used without retraining or target-domain calibration.

\paragraph{G1: same design, multi-parameter traces}
Training uses only 5.0\,V single-parameter sweeps, while G1 evaluates on 5.0\,V traces in which multiple parameters vary simultaneously.
Both AE models degrade relative to the in-domain test, especially on the anomalous class.
Nevertheless, AE-C substantially outperforms AE-R1.
This suggests that the compact feature set is less sensitive to parameter-interaction effects not present in the training distribution.

\paragraph{G2: transfer across VR designs}
On the 3.3\,V InfineonSpice design, AE-C again performs best, indicating that moderate cross-design transfer is possible when the reference voltage and stabilization dynamics remain sufficiently close to the training domain.
In contrast, both models perform poorly on the 1.5\,V design.
The main failure mode is a collapse of normal-class F1, meaning that many normal samples are classified as anomalous.
This indicates a substantial shift in normal stabilization behavior that is not compensated by the current normalization and feature set.

\paragraph{G3: simulator-to-simulator transfer}
Transfer from InfineonSpice to transistor-level simulation is the most challenging setting.
The transistor-level simulator changes the distribution of normal stabilization features enough that models trained only on InfineonSpice normals are poorly calibrated.
This supports the need for lightweight target-domain calibration, for example threshold recalibration using a small set of transistor-level normal samples, or feature-distribution alignment between simulators.

\textbf{RQ2 has therefore a mixed answer.
\approach learned predicates transfer reasonably to multi-parameter traces of the same design and to the nearby 3.3\,V design, especially when using the compact feature set.
However, large design shifts and simulator shifts require additional calibration.}

\subsection{RQ3: Symbolic-only Predicate}
\label{sec:ablations}

To assess the contribution of learned predicates, we compare \approach with a purely symbolic baseline inspired by~\cite{DBLP:conf/cav/FerrereMNU15}.
The baseline uses the same regular expression to identify perturbation intervals, but replaces the learned stabilization analysis by a fixed subinterval, namely the last 30\% of each perturbation interval.
On this subinterval, it computes three simple measurements: the average signal value, the difference between the maximum and minimum sample values, and the maximum deviation from the target value.
Thresholds are calibrated from 115 labeled traces by partitioning them into normal and anomalous sets, computing the class-wise average of each measurement, and setting each threshold to the midpoint between the corresponding normal and anomalous averages.
The resulting symbolic condition classifies an interval as normal iff all three measurements remain below their thresholds, and as anomalous otherwise.
The remaining 66 traces yield 230 valid intervals for evaluation.
Table~\ref{tab:baseline-model} reports the resulting metrics, computed with respect to the anomaly class (nok).

\begin{table}[t]
\centering
\caption{Symbolic-only baseline on the 5.0\,V behavioral model.}
\label{tab:baseline-model}
\begin{adjustbox}{max width=\linewidth}
\begin{tabular}{l c c c c c}
\toprule
\textbf{Predicate} & \textbf{Acc.} & \textbf{F1(ok)} & \textbf{F1(nok)} &
\textbf{P(nok)} & \textbf{R(nok)} \\
\midrule
Symbolic-only &
0.774 & 0.863 & 0.350 & 0.438 & 0.292 \\
\bottomrule
\end{tabular}
\end{adjustbox}
\end{table}

The symbolic baseline achieves a reasonable normal-class performance but misses most anomalies: only 14 of 48 anomalous intervals are detected.
Its F1(nok) of 0.350 is far below the learned predicates in Table~\ref{tab:ae-models}.

\textbf{This gives a positive answer to RQ3, confirming the intuition that learned predicates provide a significant advantage over 
this purely symbolic threshold-based predicate 
for these types of signal properties.}

\subsection{RQ4: Robustness to ML Methods}
\label{sec:learned-predicate-comparison}

To assess how robust \approach is to different machine learning algorithms for predicting anomalies, we compare the AE predicate with an OCSVM predicate while keeping the \speclang interval detection, stabilization localization, feature extraction, data splits, and normalization fixed.
The comparison includes two controlled OCSVM variants, OCSVM(AE-R1) and OCSVM(AE-C), which use exactly the same feature sets as AE-R1 and AE-C; this isolates the effect of replacing the learned predicate.
We also report OCSVM-B, the best OCSVM-selected feature subset found by the same nested cross-validation protocol.

Table~\ref{tab:learned-predicate-ablation} reports the learned-predicate comparison across the in-domain test set and the three generalization scenarios.
Each table entry reports three values in the order accuracy/F1(ok)/F1(nok).
Including all scenarios is important: the in-domain test alone would suggest that OCSVM is preferable, whereas the transfer settings reveal that the choice of learned predicate does not remove the dominant distribution shifts.

\begin{table*}[t]
\centering
\caption{Comparison of learned predicates. Each entry is accuracy/F1(ok)/F1(nok). OCSVM(AE-R1) and OCSVM(AE-C) keep the AE-selected feature sets fixed; OCSVM-B uses the best OCSVM-selected four-feature subset \{slope, FFT peak-to-mean ratio, FFT spectral centroid, FFT band energy\}.}
\label{tab:learned-predicate-ablation}
\scriptsize
\setlength{\tabcolsep}{3pt}
\begin{adjustbox}{max width=\textwidth}
\begin{tabular}{l c c c c c}
\toprule
\textbf{Scenario} & \textbf{AE-R1} & \textbf{OCSVM(AE-R1)} & \textbf{AE-C} & \textbf{OCSVM(AE-C)} & \textbf{OCSVM-B} \\
\midrule
In-domain 5.0\,V &
$0.878/0.877/0.879$ &
$0.891/0.884/0.898$ &
$0.870/0.867/0.872$ &
$0.891/0.887/0.895$ &
$0.900/0.895/0.905$ \\

G1: same design, multi-param. &
$0.667/0.758/0.467$ &
$0.646/0.712/0.541$ &
$0.771/0.845/0.560$ &
$0.667/0.742/0.529$ &
$0.729/0.800/0.581$ \\

G2: 3.3\,V InfineonSpice &
$0.641/0.720/0.500$ &
$0.513/0.578/0.424$ &
$0.769/0.836/0.609$ &
$0.667/0.745/0.519$ &
$0.615/0.694/0.483$ \\

G2: 1.5\,V InfineonSpice &
$0.366/0.071/0.519$ &
$0.342/0.000/0.509$ &
$0.439/0.258/0.549$ &
$0.342/0.000/0.509$ &
$0.463/0.313/0.560$ \\

G3: transistor-level &
$0.266/0.319/0.203$ &
$0.109/0.034/0.174$ &
$0.453/0.615/0.054$ &
$0.344/0.462/0.160$ &
$0.172/0.159/0.185$ \\
\bottomrule
\end{tabular}
\end{adjustbox}
\end{table*}

The OCSVM-selected configuration, denoted OCSVM-B, uses four features: slope, FFT peak-to-mean ratio, FFT spectral centroid, and FFT band energy.
Thus, the best OCSVM model is already compact and comparable in size to the compact AE feature set.
Interestingly, the compact AE and the best OCSVM feature sets both select slope, spectral centroid, and band energy, suggesting that these measurements are robust descriptors of stabilization behavior across one-class predicates.
They differ in the fourth feature: the AE selects the stabilization offset, whereas the OCSVM selects the FFT peak-to-mean ratio.

\textbf{The answer to RQ4 is therefore nuanced.
In-domain, OCSVM is highly competitive: with its own selected four-feature subset, it achieves the best accuracy and F1 scores.
This shows that \speclang-derived features are not tied to a particular neural architecture.
However, OCSVM does not systematically improve generalization.
AE-C is strongest on G1 in accuracy and normal-class F1, best on all reported metrics for the 3.3\,V design in G2, and most effective at preserving normal-class behavior in G3.
On the 1.5\,V design, all models perform poorly, indicating that the main limitation is domain shift rather than the choice between AE and OCSVM.}

\subsection{Threats to Validity}
\label{sec:threats-and-implications}

\paragraph{Internal validity}
The evaluation relies on expert-provided labels. Hence, the reported performance metrics may be sensitive to labeling noise. In addition, the one-class models are trained exclusively on normal samples, while labeled anomalies are used during cross-validation to rank feature subsets. This may introduce a mild bias toward the observed anomaly types. Future work should evaluate feature selection based only on normal samples or expert-defined criteria.

\paragraph{Construct validity}
Our approach assumes that stabilization intervals can be identified through \speclang. However, a non-negligible fraction of candidate intervals is discarded because the piece-wise linearization procedure fails to identify a valid stabilization regime. Consequently, the evaluation may not fully capture behaviors that fall outside the assumptions of the segmentation operator. Improving the robustness of the interval extraction procedure is therefore an important direction for future work.

\paragraph{External validity}
The empirical evaluation was conducted on a single industrial case study involving a voltage regulator. Although the stabilization and oscillation phenomena considered in this work are representative of common challenges in analog circuit design, the generalizability of the results to other analog or mixed-signal systems remains to be demonstrated. Furthermore, the observed performance degradation on the 1.5\,V design and transistor-level traces indicates that domain shift between source and target configurations can substantially affect detection accuracy. Practical deployment may therefore require target-domain calibration, such as threshold adaptation using a small number of normal target-domain samples or feature-wise normalization across domains.

\paragraph{Conclusion validity}
Several evaluations, particularly the out-of-domain experiments, rely on relatively small expert-labeled test sets. The reported results should therefore be interpreted as indicative trends rather than statistically definitive estimates. This limitation is especially relevant for anomaly-detection performance, where only a limited number of anomalous traces are available.

\section{Conclusion and Future Work}
\label{sec:future_work}
In this paper, we presented \approach, an approach that combines formal signal analysis with data-driven anomaly detection for analog and hybrid systems, showing the synergies between formal methods and machine learning for CPS monitoring workflows. We demonstrated on an industrial voltage regulator case study that \approach can detect subtle stabilization anomalies that are difficult to characterize using purely symbolic specifications. The experiments show that \approach is robust under moderate distribution shifts, including multi-parameter variations and transfer across regulator designs, and that it is not tied to a particular learning architecture. 

Our framework can be extended in several directions. For instance, when one domain is used for learning and another for evaluation, there may be systematic shifts between the training and evaluation domains. It would be interesting to see if these can be accounted for in the specification or machine learning. This paper is intended to propose a framework rather than a definitive set of Measurement Operators; the set chosen by us can certainly be improved. Finally, our methodology can be integrated into a verification loop, driving the selection of simulation parameters toward areas in which the system may exhibit worse performance.


\FloatBarrier 
\bibliography{main}

@inproceedings{DBLP:conf/bigdataconf/AhmadSNK20,
  author       = {Sabtain Ahmad and
                  Kevin Styp{-}Rekowski and
                  Sasho Nedelkoski and
                  Odej Kao},
  editor       = {Xintao Wu and
                  Chris Jermaine and
                  Li Xiong and
                  Xiaohua Hu and
                  Olivera Kotevska and
                  Siyuan Lu and
                  Weija Xu and
                  Srinivas Aluru and
                  Chengxiang Zhai and
                  Eyhab Al{-}Masri and
                  Zhiyuan Chen and
                  Jeff Saltz},
  title        = {Autoencoder-based Condition Monitoring and Anomaly Detection Method
                  for Rotating Machines},
  booktitle    = {2020 {IEEE} International Conference on Big Data {(IEEE} BigData 2020),
                  Atlanta, GA, USA, December 10-13, 2020},
  pages        = {4093--4102},
  publisher    = {{IEEE}},
  year         = {2020},
  url          = {https://doi.org/10.1109/BigData50022.2020.9378015},
  doi          = {10.1109/BIGDATA50022.2020.9378015},
  bibsource    = {dblp computer science bibliography, https://dblp.org}
}

@inproceedings{ozay2016exact,
  title={An exact and efficient algorithm for segmentation of ARX models},
  author={Ozay, Necmiye},
  booktitle={2016 American Control Conference (ACC)},
  pages={38--41},
  year={2016},
  organization={IEEE}
}

@inproceedings{DBLP:conf/cav/FerrereMNU15,
  author       = {Thomas Ferr{\`{e}}re and
                  Oded Maler and
                  Dejan Nickovic and
                  Dogan Ulus},
  editor       = {Daniel Kroening and
                  Corina S. Pasareanu},
  title        = {Measuring with Timed Patterns},
  booktitle    = {Computer Aided Verification - 27th International Conference, {CAV}
                  2015, San Francisco, CA, USA, July 18-24, 2015, Proceedings, Part
                  {II}},
  series       = {Lecture Notes in Computer Science},
  volume       = {9207},
  pages        = {322--337},
  publisher    = {Springer},
  year         = {2015},
  url          = {https://doi.org/10.1007/978-3-319-21668-3\_19},
  doi          = {10.1007/978-3-319-21668-3\_19},
  bibsource    = {dblp computer science bibliography, https://dblp.org}
}

@inproceedings{DBLP:conf/formats/BakhirkinFMU17,
  author       = {Alexey Bakhirkin and
                  Thomas Ferr{\`{e}}re and
                  Oded Maler and
                  Dogan Ulus},
  editor       = {Alessandro Abate and
                  Gilles Geeraerts},
  title        = {On the Quantitative Semantics of Regular Expressions over Real-Valued
                  Signals},
  booktitle    = {Formal Modeling and Analysis of Timed Systems - 15th International
                  Conference, {FORMATS} 2017, Berlin, Germany, September 5-7, 2017,
                  Proceedings},
  series       = {Lecture Notes in Computer Science},
  pages        = {189--206},
  publisher    = {Springer},
  year         = {2017},
  url          = {https://doi.org/10.1007/978-3-319-65765-3\_11},
  doi          = {10.1007/978-3-319-65765-3\_11},
  bibsource    = {dblp computer science bibliography, https://dblp.org}
}

@inproceedings{DBLP:conf/formats/MalerN04,
  author       = {Oded Maler and
                  Dejan Nickovic},
  editor       = {Yassine Lakhnech and
                  Sergio Yovine},
  title        = {Monitoring Temporal Properties of Continuous Signals},
  booktitle    = {Formal Techniques, Modelling and Analysis of Timed and Fault-Tolerant
                  Systems, Joint International Conferences on Formal Modelling and Analysis
                  of Timed Systems, {FORMATS} 2004 and Formal Techniques in Real-Time
                  and Fault-Tolerant Systems, {FTRTFT} 2004, Grenoble, France, September
                  22-24, 2004, Proceedings},
  series       = {Lecture Notes in Computer Science},
  pages        = {152--166},
  publisher    = {Springer},
  year         = {2004},
  url          = {https://doi.org/10.1007/978-3-540-30206-3\_12},
  doi          = {10.1007/978-3-540-30206-3\_12},
  bibsource    = {dblp computer science bibliography, https://dblp.org}
}

@inproceedings{DBLP:conf/formats/UlusFAM14,
  author       = {Dogan Ulus and
                  Thomas Ferr{\`{e}}re and
                  Eugene Asarin and
                  Oded Maler},
  editor       = {Axel Legay and
                  Marius Bozga},
  title        = {Timed Pattern Matching},
  booktitle    = {Formal Modeling and Analysis of Timed Systems - 12th International
                  Conference, {FORMATS} 2014, Florence, Italy, September 8-10, 2014.
                  Proceedings},
  series       = {Lecture Notes in Computer Science},
  pages        = {222--236},
  publisher    = {Springer},
  year         = {2014},
  url          = {https://doi.org/10.1007/978-3-319-10512-3\_16},
  doi          = {10.1007/978-3-319-10512-3\_16},
  bibsource    = {dblp computer science bibliography, https://dblp.org}
}

@inproceedings{DBLP:conf/kdd/HundmanCLCS18,
  author       = {Kyle Hundman and
                  Valentino Constantinou and
                  Christopher Laporte and
                  Ian Colwell and
                  Tom S{\"{o}}derstr{\"{o}}m},
  editor       = {Yike Guo and
                  Faisal Farooq},
  title        = {Detecting Spacecraft Anomalies Using LSTMs and Nonparametric Dynamic
                  Thresholding},
  booktitle    = {Proceedings of the 24th {ACM} {SIGKDD} International Conference on
                  Knowledge Discovery {\&} Data Mining, {KDD} 2018, London, UK,
                  August 19-23, 2018},
  pages        = {387--395},
  publisher    = {{ACM}},
  year         = {2018},
  url          = {https://doi.org/10.1145/3219819.3219845},
  doi          = {10.1145/3219819.3219845},
  bibsource    = {dblp computer science bibliography, https://dblp.org}
}

@article{DBLP:journals/tac/KongJB17,
  author       = {Zhaodan Kong and
                  Austin Jones and
                  Calin Belta},
  title        = {Temporal Logics for Learning and Detection of Anomalous Behavior},
  journal      = {{IEEE} Trans. Autom. Control.},
  volume       = {62},
  number       = {3},
  pages        = {1210--1222},
  year         = {2017},
  url          = {https://doi.org/10.1109/TAC.2016.2585083},
  doi          = {10.1109/TAC.2016.2585083},
  bibsource    = {dblp computer science bibliography, https://dblp.org}
}

@article{DBLP:journals/fmsd/DeshmukhDGJJS17,
  author       = {Jyotirmoy V. Deshmukh and
                  Alexandre Donz{\'{e}} and
                  Shromona Ghosh and
                  Xiaoqing Jin and
                  Garvit Juniwal and
                  Sanjit A. Seshia},
  title        = {Robust online monitoring of signal temporal logic},
  journal      = {Formal Methods Syst. Des.},
  volume       = {51},
  number       = {1},
  pages        = {5--30},
  year         = {2017},
  url          = {https://doi.org/10.1007/s10703-017-0286-7},
  doi          = {10.1007/S10703-017-0286-7},
  bibsource    = {dblp computer science bibliography, https://dblp.org}
}

@article{DBLP:journals/jacm/AsarinCM02,
  author       = {Eugene Asarin and
                  Paul Caspi and
                  Oded Maler},
  title        = {Timed regular expressions},
  journal      = {J. {ACM}},
  volume       = {49},
  number       = {2},
  pages        = {172--206},
  year         = {2002},
  url          = {https://doi.org/10.1145/506147.506151},
  doi          = {10.1145/506147.506151},
  bibsource    = {dblp computer science bibliography, https://dblp.org}
}

@article{DBLP:journals/tcs/FainekosP09,
  author       = {Georgios E. Fainekos and
                  George J. Pappas},
  title        = {Robustness of temporal logic specifications for continuous-time signals},
  journal      = {Theor. Comput. Sci.},
  volume       = {410},
  number       = {42},
  pages        = {4262--4291},
  year         = {2009},
  url          = {https://doi.org/10.1016/j.tcs.2009.06.021},
  doi          = {10.1016/J.TCS.2009.06.021},
  bibsource    = {dblp computer science bibliography, https://dblp.org}
}

@inproceedings{DBLP:conf/icdm/YehZUBDDSMK16,
  author       = {Chin{-}Chia Michael Yeh and
                  Yan Zhu and
                  Liudmila Ulanova and
                  Nurjahan Begum and
                  Yifei Ding and
                  Hoang Anh Dau and
                  Diego Furtado Silva and
                  Abdullah Mueen and
                  Eamonn J. Keogh},
  editor       = {Francesco Bonchi and
                  Josep Domingo{-}Ferrer and
                  Ricardo Baeza{-}Yates and
                  Zhi{-}Hua Zhou and
                  Xindong Wu},
  title        = {Matrix Profile {I:} All Pairs Similarity Joins for Time Series: {A}
                  Unifying View That Includes Motifs, Discords and Shapelets},
  booktitle    = {{IEEE} 16th International Conference on Data Mining, {ICDM} 2016,
                  December 12-15, 2016, Barcelona, Spain},
  pages        = {1317--1322},
  publisher    = {{IEEE} Computer Society},
  year         = {2016},
  url          = {https://doi.org/10.1109/ICDM.2016.0179},
  doi          = {10.1109/ICDM.2016.0179},
  bibsource    = {dblp computer science bibliography, https://dblp.org}
}

@inproceedings{DBLP:conf/pricai/SakuradaY14,
  author       = {Mayu Sakurada and
                  Takehisa Yairi},
  editor       = {Ashfaqur Rahman and
                  Jeremiah D. Deng and
                  Jiuyong Li},
  title        = {Anomaly Detection Using Autoencoders with Nonlinear Dimensionality
                  Reduction},
  booktitle    = {Proceedings of the {MLSDA} 2014 2nd Workshop on Machine Learning for
                  Sensory Data Analysis, Gold Coast, Australia, QLD, Australia, December
                  2, 2014},
  pages        = {4},
  publisher    = {{ACM}},
  year         = {2014},
  url          = {https://doi.org/10.1145/2689746.2689747},
  doi          = {10.1145/2689746.2689747},
  bibsource    = {dblp computer science bibliography, https://dblp.org}
}

@inproceedings{DBLP:conf/nips/ScholkopfWSSP99,
  author       = {Bernhard Sch{\"{o}}lkopf and
                  Robert C. Williamson and
                  Alexander J. Smola and
                  John Shawe{-}Taylor and
                  John C. Platt},
  editor       = {Sara A. Solla and
                  Todd K. Leen and
                  Klaus{-}Robert M{\"{u}}ller},
  title        = {Support Vector Method for Novelty Detection},
  booktitle    = {Advances in Neural Information Processing Systems 12, {[NIPS} Conference,
                  Denver, Colorado, USA, November 29 - December 4, 1999]},
  pages        = {582--588},
  publisher    = {The {MIT} Press},
  year         = {1999},
  url          = {http://papers.nips.cc/paper/1723-support-vector-method-for-novelty-detection},
  bibsource    = {dblp computer science bibliography, https://dblp.org}
}

@article{Esmaei23,
    author = {Fatemeh Esmaeili and Erica Cassie and Hong Phan T. Nguyen and Natalie O. V. Plank and
Charles P. Unsworth and Alan Wang},
    title = {Anomaly Detection for Sensor Signals Utilizing Deep Learning
Autoencoder-Based Neural Networks},
    journal = {Bioengineering},
    year = 2023
}

@techreport{ti1999ldo,
  author      = {Lee, Bang S.},
  title       = {Understanding the Terms and Definitions of LDO Voltage Regulators},
  institution = {Texas Instruments},
  number      = {SLVA079},
  year        = {1999},
  month       = oct,
  type        = {Application Report}
}

@misc{infineonspice,
  author       = {{Infineon Technologies AG}},
  title        = {{InfineonSpice Offline Simulation Tool}},
  howpublished = {\url{https://www.infineon.com/design-resources/simulation-modeling/infineonspice-offline-simulation-tool}},
  note         = {Accessed: 2026-05-04}
}
\end{document}